\documentclass[default]{sn-jnl}

\usepackage{graphicx}%
\usepackage{multirow}%
\usepackage{amsmath,amssymb,amsfonts}%
\usepackage{amsthm}%
\usepackage{mathrsfs}%
\usepackage[title]{appendix}%
\usepackage{xcolor}%
\usepackage{textcomp}%
\usepackage{manyfoot}%
\usepackage{booktabs}%
\usepackage{algorithm}%
\usepackage{algorithmicx}%
\usepackage{algpseudocode}%
\usepackage{listings}%

\theoremstyle{thmstyleone}%
\theoremstyle{thmstyletwo}%

\theoremstyle{thmstylethree}%
\usepackage{natbib}
\begin{document}

\title[Article Title]{GCAM: Gaussian and causal-attention model of food fine-grained recognition}


\author*[1]{\fnm{Guohang} \sur{Zhuang}}\email{zoralastrapelft@gmail.com}

\author[2]{\fnm{Yue} \sur{Hu}}\email{3236305310@qq.com}
\equalcont{These authors contributed equally to this work.}

\author[3]{\fnm{Tianxing} \sur{Yan}}\email{1394698398@qq.com}

\author[4]{\fnm{Jiazhan} \sur{Gao}}\email{gjz6611@stu.xju.edu.cn}


\affil*[1]{\orgdiv{School of Computer and Information}, \orgname{Hefei University of Technology}, \orgaddress{\street{Shushan District}, \city{Hefei}, \postcode{230009}, \state{Anhui}, \country{China}}}

\affil[2]{\orgdiv{School of Food Science and Engineering}, \orgname{Northern University for Nationalities}, \orgaddress{\street{Xixia District}, \city{Yinchuan}, \postcode{750030}, \state{Ningxia}, \country{China}}}

\affil[3]{\orgdiv{School of Electrical Engineering}, \orgname{Xinjiang University}, \orgaddress{\street{Huarui Street, Shuimogou District}, \city{Urumqi City}, \postcode{830046}, \state{Xinjiang Uygur Autonomous Region}, \country{China}}}

\affil[4]{\orgdiv{school of computer science and technology}, \orgname{Xinjiang University}, \orgaddress{\street{Huarui Street, Shuimogou District}, \city{Urumqi City}, \postcode{830046}, \state{Xinjiang Uygur Autonomous Region}, \country{China}}}


\abstract{Currently, most food recognition relies on deep learning for category classification. However, these approaches struggle to effectively distinguish between visually similar food samples, highlighting the pressing need to address fine-grained issues in food recognition. To mitigate these challenges, we propose the adoption of a Gaussian and causal-attention model for fine-grained object recognition.In particular, we train to obtain Gaussian features over target regions, followed by the extraction of fine-grained features from the objects, thereby enhancing the feature mapping capabilities of the target regions. To counteract data drift resulting from uneven data distributions, we employ a counterfactual reasoning approach. By using counterfactual interventions, we analyze the impact of the learned image attention mechanism on network predictions, enabling the network to acquire more useful attention weights for fine-grained image recognition. Finally, we design a learnable loss strategy to balance training stability across various modules, ultimately improving the accuracy of the final target recognition. We validate our approach on four relevant datasets, demonstrating its excellent performance across these four datasets.We experimentally show that GCAM surpasses state-of-the-art methods on the ETH-FOOD101, UECFOOD256, and Vireo-FOOD172 datasets. Furthermore, our approach also achieves state-of-the-art performance on the CUB-200 dataset}

\keywords{Gaussian function, Counterfactuals are inferences, Fine-grained identification of food, attention mechanism}



\maketitle

\section{Introduction}\label{sec1}
\begin{figure}
\centering 
\includegraphics[width=1.0\columnwidth]{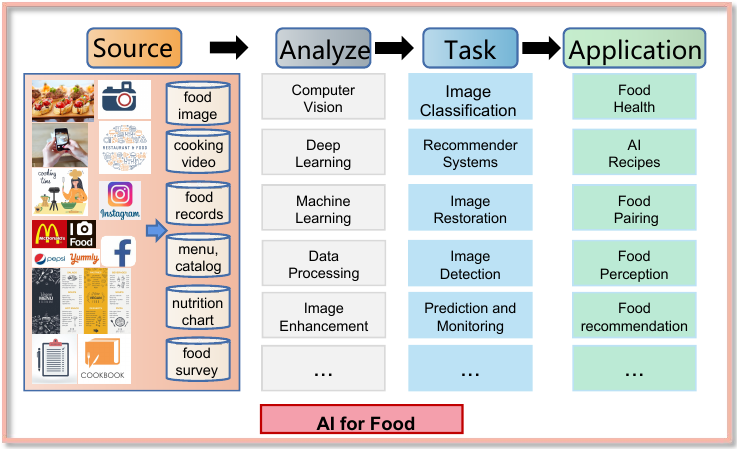}
\caption{food system}
\label{fig1}
\end{figure}

Food is of great importance to human life and serves as the foundation for human survival. Research related to food extends into various societal domains, encompassing not only food testing but also supporting the application of devices to guide human behavior, improve human health, and understand culinary cultures, among other services\cite{1,2}. Among these, food recognition plays a crucial role in large-scale food data, involving the use of computer vision and machine learning techniques to automatically identify and categorize food images. This technology finds applications in various fields and holds significant value and impact, such as in the food industry, where the identification of food image information helps enhance consumer engagement and boost food sales\cite{3}. As shown in Fig \ref{fig1}, the fusion of food and artificial intelligence fields can be advantageous for improving the daily lives of humanity.

\begin{figure}
\centering 
\includegraphics[width=0.6\columnwidth]{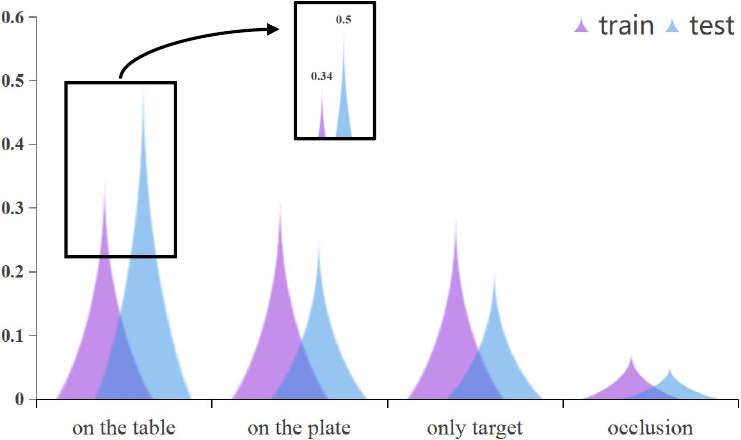}
\caption{Distribution of food categories by location}
\label{fig2}
\end{figure}

Food varieties are typically diverse, and foods within the same category often exhibit varying appearances and colors, posing significant challenges for food recognition. This significantly reduces the reliability of local feature descriptions in networks and the performance of traditional methods based on local features. Complex spatial relationships also diminish the reliability of optimal feature extraction methods. Typically, researchers categorize this problem as fine-grained recognition, which holds important research value in fine-grained visual recognition of food\cite{4}. Fine-grained image recognition refers to the recognition of different sub-classes within the same category of objects, such as identifying different types of cakes or noodles. In \cite{5} research, it has been observed that many common objects (like giraffes or boats) can be robustly classified using models that explicitly encode spatial relationships between parts. In contrast, food exhibits weaker spatial structural characteristics, indicating that food recognition is more challenging than recognizing regular objects. Additionally, food samples follow a long-tailed distribution\cite{CAL}, which results in food images with a small sample size not receiving adequate attention during network training, affecting the performance of food recognition. As shown in Fig \ref{fig2}, the distribution of the same food category is uneven.

In this work, we propose a \textbf{G}aussian and \textbf{C}ausal-\textbf{A}ttention \textbf{M}odel for fine-grained food recognition as a solution to the challenges at hand. This approach presents a straightforward yet effective framework for food identification. GCAM is a two-stage network that takes both the original image (representing coarse granularity) and the attention-cropped image (representing fine granularity) features as inputs. To obtain the fine-grained representation of the image, we introduce the Feature Gaussian Fusion (FGF) module. Within FGF, we employ network-based learning of food distribution functions, integrating them into the original image through weighted fusion, thereby enabling the network to acquire superior feature representations during the training process. The refined image is segmented from the attention weight map generated. Furthermore, leveraging counterfactual causal relationships, we quantify the quality of attention by comparing the impact of actual and counterfactual scenarios on the final prediction. To enhance the training process, we devise an effective loss learning strategy. This paper provides an in-depth exposition of experimental results and comparative analyses. Our contributions encompass the following:

(1) A novel network, GCAM, is introduced for food recognition. In FGF, through network-based learning, models the Gaussian distribution function of the object's spatial regions within the image. This distribution assists in mitigating the influence of background elements on predictions during the feature extraction process. 

(2) In response to the issue of instability in training effectiveness associated with attention maps, we establish a causal graph model to optimize attention maps, resulting in significant improvements in training outcomes. Furthermore, we devise an effective Loss Learning Strategy (LLS) to enhance network training. 

(3) Our proposed algorithm exhibits state-of-the-art performance on datasets including Vireo-FOOD172, FOOD256, and Food-101. Additionally, to validate the network's effectiveness on other fine-grained datasets, we conducted training and testing on CUB-200, achieving optimal accuracy.
\section{related work}\label{sec2}

We briefly discuss relevant literature for this method, encompassing three research areas: 1) Food Image Recognition, 2) Fine-grained Image Recognition.

\textbf{Food Image Recognition}: Food image recognition algorithms have been pivotal and central in various food-related tasks such as dietary assessment, food perception, and food recommendation\cite{6}. Categorizing food image recognition methods based on feature types, we find two primary categories: recognition based on handcrafted features and recognition based on deep features. Classical handcrafted feature methods include the use of operators like LBP\cite{7}, SIFT\cite{8}, and HOG\cite{9}. As deep learning has demonstrated its effectiveness across various applications, traditional feature extraction methods have been largely replaced by deep networks. Most existing literature on food recognition employs a standard CNN network structure fine-tuned on food datasets, yielding satisfactory results.

Kagaya et al.\cite{10} employed the AlexNet network to extract image features for food recognition, while Ming et al.\cite{11} utilized the Resnet50 network for food identification. Steinbrener et al.\cite{12} employed an improved GoogLeNet for the classification of fruits and vegetables. These methods directly extract features based on CNNs, often failing to consider the distinctive characteristics of food, introducing irrelevant noise into the network.

\begin{figure*}[!t]
\centering
\includegraphics[width=\linewidth]{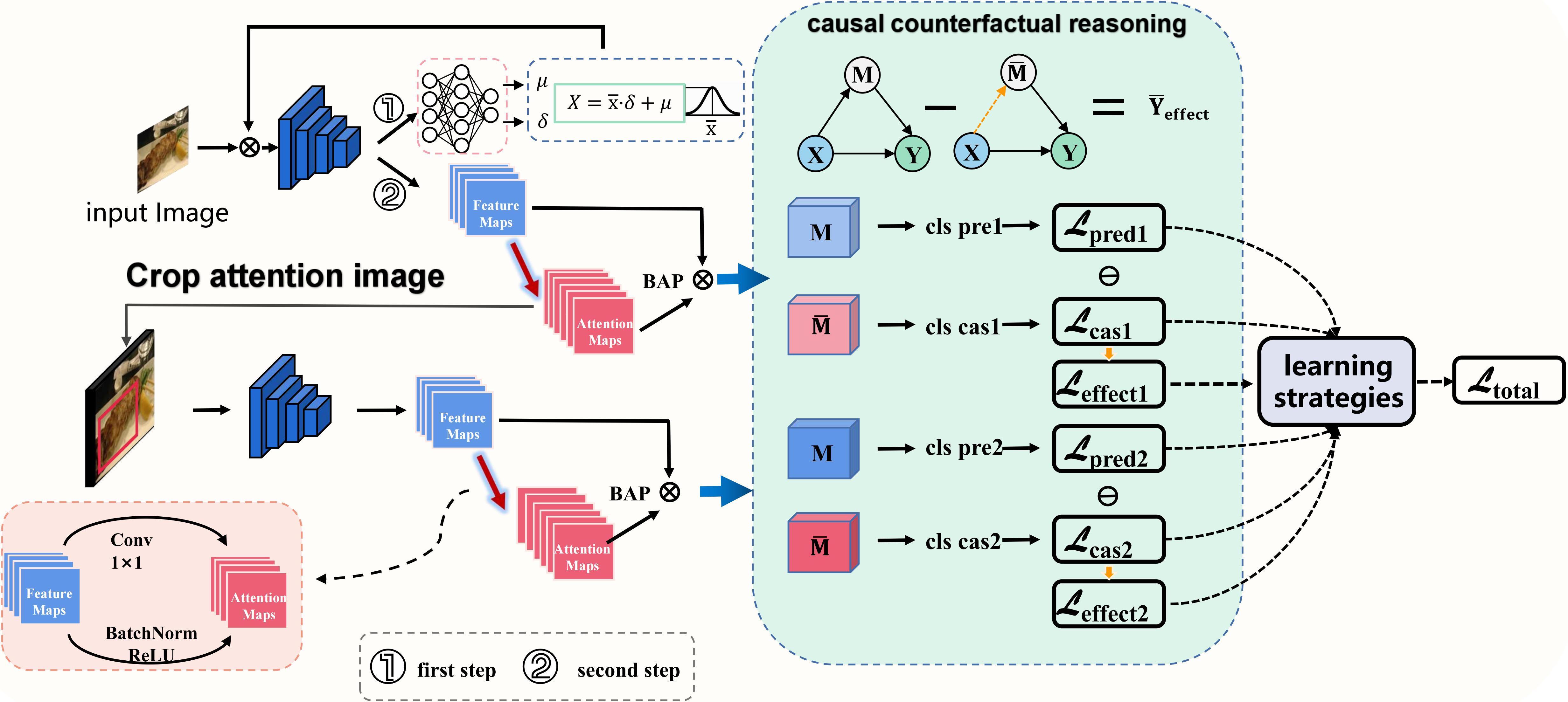}
\caption{Overall architecture of GCAM for food identification. Input the original image, obtain the coarse-grained features through FGF, and then obtain the refined image through clipping. Among them, CRA improves the quality of the network attention mechanism and reduces the global discriminative ability of the network attention mechanism. The final losses are pooled into LLS for optimization.}
\label{fig3}
\end{figure*}

In recent years, some researchers have begun designing deep network structures tailored for food images. Martinel et al.\cite{13} proposed the WISeR network, consisting of two branches: a Wide Residual Networks (WRN) branch for extracting general visual features of food images and a Slice Convolutional Networks branch for capturing the vertical structure of food items (e.g., burgers, pizzas). The features from these two branches are concatenated to obtain the final feature representation. Thanks to the fusion of global features and vertical structural features of food images, WISeR achieved optimal recognition performance on multiple datasets at that time. Chen et al.\cite{14} analyzed the differences between global and local region-based image understanding in ingredient recognition and explored the impact of various learning methods, including single-task and multi-task learning, on ingredient recognition.

While these methods proposed network architectures to enhance food image perception, they struggle when backgrounds and food items closely resemble each other, making it difficult to represent features distinctly among similar food items.

\textbf{Fine-grained Image Recognition}: Fine-grained visual classification (FGVC) is a method often employed to address "intra-class classification" and has seen significant development in the field of computer vision \cite{15,16,b1}. Hou et al.\cite{17} introduced the VegFru dataset for fine-grained image classification of fruits and vegetables, along with the HybridNet framework composed of two CNNs capable of extracting features at multiple granularities for coarse and fine-grained labels. These features are then fused through bilinear pooling and similar techniques to obtain the final feature representation. Hu et al.\cite{18} proposed an approach for fine-grained visual classification tasks, combining weakly supervised learning with image enhancement methods and attention mechanisms to enable the network to focus on the "informative" regions within images without requiring additional labeling information, achieving state-of-the-art performance in fine-grained classification tasks.

However, these methods, while effective in capturing differences within classes, often concentrate on prominent differences in feature maps, overlooking subtle inter-class differences. It is essential to guide the network to pay attention to distinctions among classes that are more prone to confusion, essentially identifying which parts of an object the network focuses on.

\section{Network}\label{sec3}
Fig \ref{fig3} illustrates the framework of the proposed GCAM network. It takes an image as input, initially passing through the Swim-transformer\cite{19} as the backbone network to extract image features. Following this, there are two branches. The first branch performs coarse-grained feature extraction on the input, where the network learns a Gaussian distribution weighted by the core position of objects and fuses it back into the original image. The second branch conducts fine-grained feature extraction by employing an attention mechanism to identify regions of interest within the image, cropping the attended area from the original image, resulting in a cropped image containing the object. Subsequently, features are re-extracted, and image class prediction is performed.

Additionally, to mitigate initial attention biases during training, a counterfactual causality is employed. By quantifying the impact of factual and counterfactual scenarios on the final prediction, attention quality is measured. The network then maximizes this difference, encouraging more effective attention learning and reducing the bias from the training dataset. Furthermore, to address the issue of misleading errors due to large gradients, an effective loss learning strategy controls the entire training process, ensuring that each task begins training only after its predecessor task has been adequately trained.

\subsection{Feature Gaussian Fusion, FGF}\label{subsec2}

\begin{figure}
\centering 
\includegraphics[width=0.6\columnwidth]{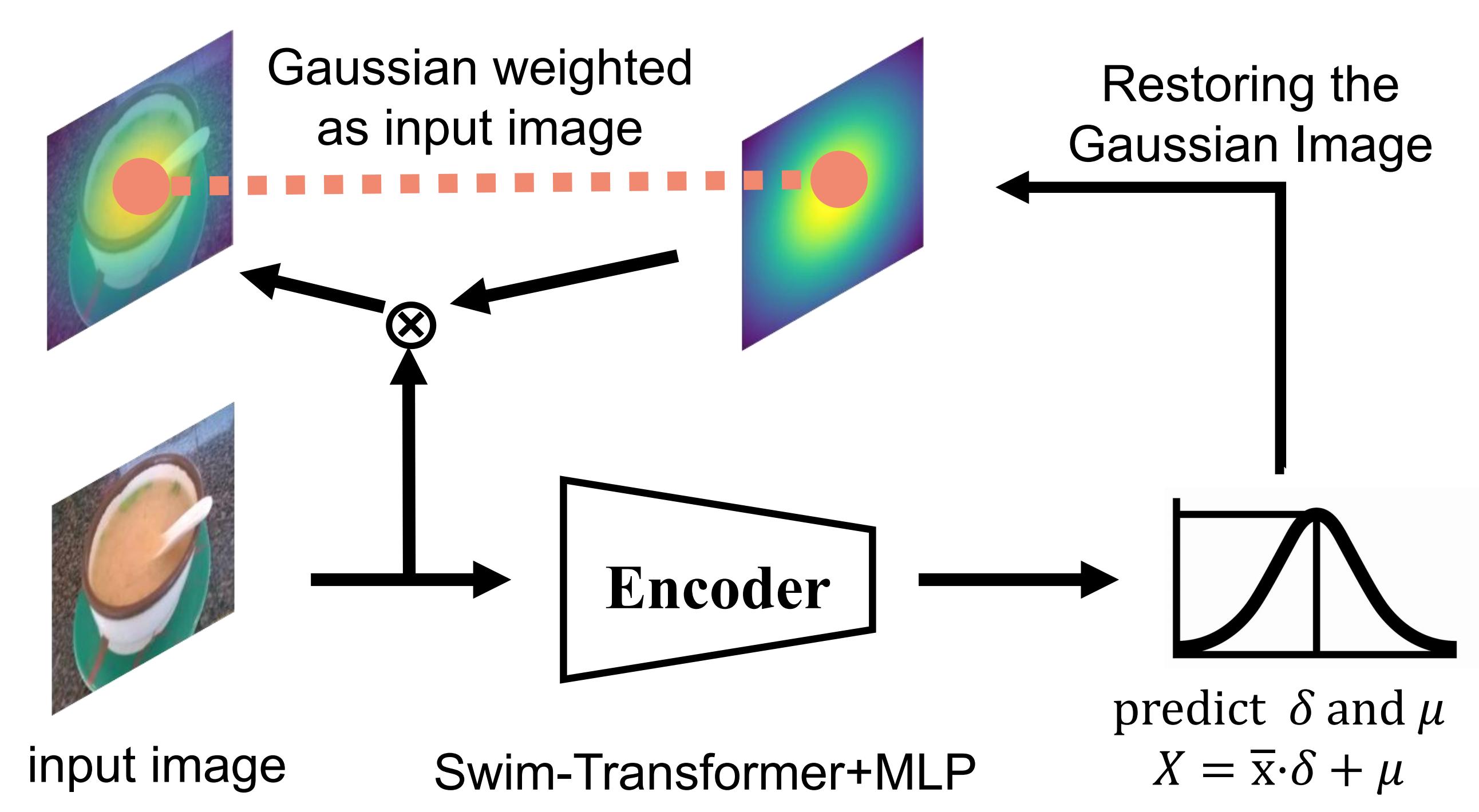}
\caption{Gaussian feature fusion process.}
\label{fig4}
\end{figure}

The distribution of objects in images typically follows a Gaussian distribution. In object detection, supervised information can be utilized through the supervised information of object center points, akin to central point heatmaps\cite{20,21}. Specifically, a Gaussian distribution is constructed around the label of the central point, generating a heatmap that serves as a supervision signal for training the network.

We introduce the FGF module to enable the network to learn the distribution of objects within the image. We train the network to output two features through an MLP, denoted as $\sigma$ and $\mu$ (as illustrated in Fig \ref{fig4}). These two features together form a Gaussian distribution, from which a weighted map for the original image is derived. This weighted map is then used for subsequent feature extraction, as shown in equation (1):

\begin{align}
{{\rm{I}}_{(x,y)}} = {{\rm{I}}_{(i,j)}} \cdot G(\mu ,\sigma )\begin{array}{*{20}{c}}
\end{array}\
\end{align}

During the early stages of training, the network may experience Gaussian distribution shifts due to the lack of supervision signals. To mitigate this, we initialize the features $\sigma$ and $\mu$ as Gaussian maps centered on the central point of the image distribution. It's worth noting that the FGF module represents the first step in the overall network architecture, with the image weighted by the original image serving as the input for the second step.

\subsection{feature and attention}\label{subsubsec2}

The attention-feature learning step is based on weakly supervised attention region learning. Firstly, the network performs feature extraction on the weighted and fused image, resulting in feature maps. Subsequently, these feature maps undergo a convolution operation with a kernel size of 1, followed by BatchNormalization and ReLU activation, yielding attention maps. In other words, the attention maps are obtained by dimensionality reduction from the original feature maps, reducing the dimensionality from C to D, where each of the D attention maps corresponds to different object locations.

\begin{figure}
\centering 
\includegraphics[width=0.6\columnwidth]{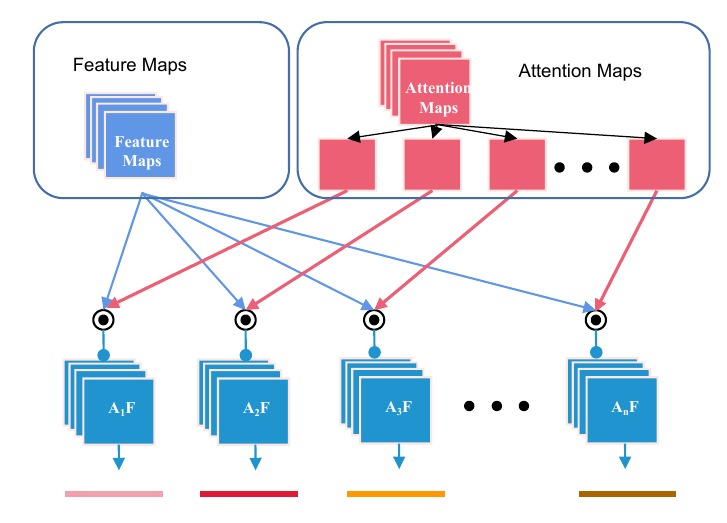}
\caption{In the image clipping process, more useful information is obtained through the interaction of feature maps and attention maps, which is used to obtain the next refinement map.}
\label{fig5}
\end{figure}

The Bilinear Attention Pooling network (BAP), as illustrated in Fig \ref{fig5} above, operates by element-wise multiplication of the feature maps with each channel of the attention map, as shown in the following equation 2:
\begin{align}
{\rm{M = }}\psi {\rm{(A,F) = }}\left( \begin{array}{l}
{a_1} \cdot F\\
{a_2} \cdot F\\
\begin{array}{*{20}{c}}
{}&{...}
\end{array}\\
{a_D} \cdot F
\end{array} \right)\
\end{align}

After multiplication, the final Feature Matrix M is obtained through pooling dimensionality reduction and splicing operations, which is the input of the final linear classification layer. Later, the network will also perform targeted enhancement on the image based on the Attention map. As shown in the following equation (3-4):

\begin{align}
{\rm{Avg}}({{\rm{A}}_D}) = \frac{1}{D}\sum\limits_1^D {{\rm{A}}{}_i} \
\end{align}
\begin{align}
{{\rm{A}}_D} = \frac{{{\rm{Avg}}({{\rm{A}}_D}) - \min ({{\rm{A}}_D})}}{{\max ({{\rm{A}}_D}) - \min ({{\rm{A}}_D})}}\
\end{align}

The extracted parts are enlarged and used as enhanced data for training. As shown in the following equation (5-6):

\begin{align}
{{\rm{I}}_p}(i,j) = \left\{ \begin{array}{l}
1\begin{array}{*{20}{c}}
{}&{}
\end{array}if{\rm{ }}{{\rm{A}}_D}(i,j) \ge 0\\
0\begin{array}{*{20}{c}}
{}&{}
\end{array}{\rm{otherwise}}
\end{array} \right.\
\end{align}

\begin{align}
{{\rm{I}}_{sed}}(x,y) = {\rm{I}}(x,y) \odot {{\rm{I}}_p}(i,j)\
\end{align}

where, ${\rm{I}}(x,y) \in {{\rm{R}}^{{\rm{W \times H \times C}}}}\ $ is the input original image. ${{\rm{I}}_{sed}}(x,y) \in {{\rm{R}}^{{\rm{W \times H \times C}}}}$\ is the subdivided input image.

\subsection{Causal counterfactual reasoning for attention, CRA}\label{subsubsec2}

Attention modules typically learn the regions of interest in an image under the supervision of the final classification loss function. However, this approach can result in a lack of a causal relationship between the predicted outcomes and the attention mechanism. Such methods also fail to teach the computer to distinguish between primary cues and biased cues. For example, as shown in Fig \ref{fig2}, in images labeled as "rice" on food labels, the background for rice in this category may include "on a plate," "on a table," "with occlusions," and "rice only." Due to uneven data distribution, such as the prevalence of "on a table," it becomes evident that attributes and environments exhibit biases. This suggests that neither the background nor individual components can be considered reliable classification cues. Consequently, the attention mechanism may incorporate knowledge of the table as part of the discriminative region for classification, limiting the model's generalization capabilities on the test set. Therefore, attention learning mechanisms do not always yield effective results, as they lack discrimination, clear definition, and robustness.

\begin{figure}
\centering 
\includegraphics[width=1.0\columnwidth]{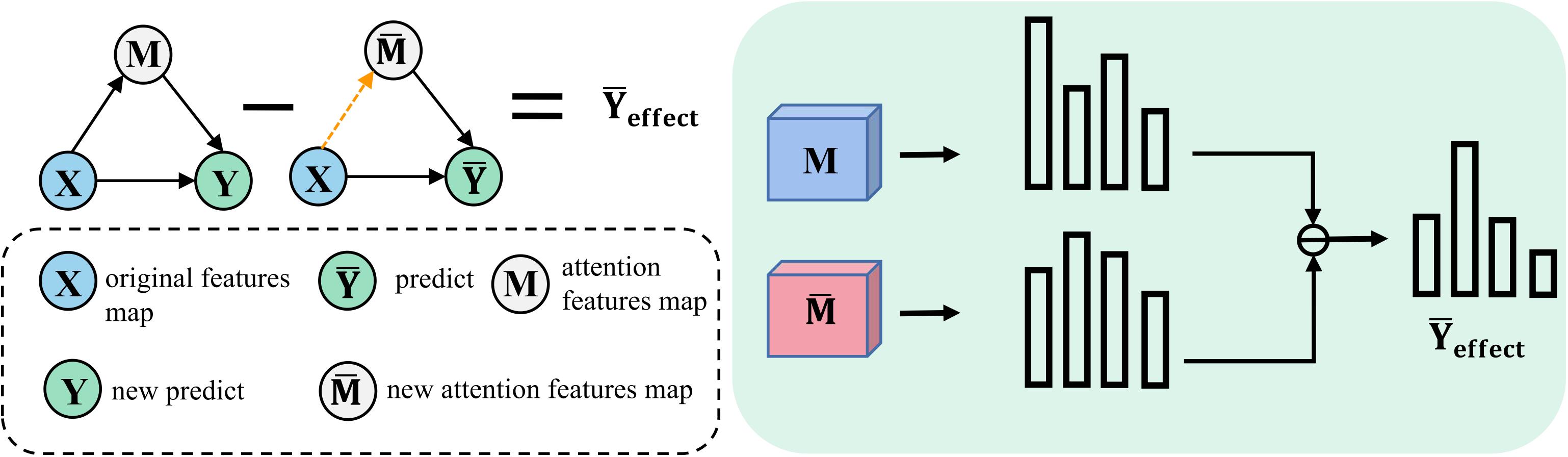}
\caption{Illustration of the CRA method. On the basis of the original attention map M, random attention ${{\rm{\bar M}}}$ is added to perform counterfactual intervention. In the next step, the counterfactual classification results are subtracted from the original classification results, the effects of the learned attention are analyzed, and they are maximized during training.}
\label{fig6}
\end{figure}

Inspired by\cite{CAL}, we combine the attention mechanism and the network to construct a causal graph, as shown in  Fig \ref{fig6}. The causal graph is also a directed acyclic graph $g\{ \ell ,\varepsilon \}$. Each variable corresponds to an $\ell$ node, and the causal $\varepsilon$ connects the relationship between these variables. 

We represent the variables of the attention module in our model through causal nodes, including the feature map (or input image) M, the learned attention map ${{\rm{\bar M}}}$, and the final prediction Y. The connectivity between nodes is denoted as ${\rm{X}} \to {\rm{M}}$, signifying that the features extracted by network X serve as input to attention map M. Additionally, input features X and M jointly influence the ultimate prediction Y, i.e., (X, ${\rm{M}} \to {\rm{X}}$). To mitigate the interference of X as an input to attention map M, we introduce an unrelated attention map denoted as (7-8):

\begin{align}
{{{\rm{\bar M}}}_I} = sign({{\rm{M}}_i}) \cdot {{\rm{M}}_i}\begin{array}{*{20}{c}}
{}&{i = 0,1,...,D}
\end{array}
\end{align}

\begin{align}
{\rm{\bar M = }}\left( \begin{array}{l}
{\rm{M}} \otimes sign({m_1})\\
{\rm{M}} \otimes sign({m_2})\\
\begin{array}{*{20}{c}}
{}&{}&{...}
\end{array}\\
{\rm{M}} \otimes sign({m_D})
\end{array} \right)
\end{align}

In this context, D represents the number of feature channels. Consequently, it is possible to substitute the learned attention map M with an unrelated attention map denoted as ${{\rm{\bar M}}}$ to ensure the preservation of feature map X. When we seek a deeper understanding of the role of a specific variable, we can eliminate all incoming connections to that variable and assign it a specific value. By approximating M, thereby subtracting ${\rm{X}} \to {\rm{M}}$, we reduce the influence of parent node X on variable M, ultimately arriving at the final prediction Y. As shown in the following equation (9-11):

\begin{align}
{\rm{Y(M = \textbf{M},X = \textbf{X}) = }}C{\rm{([f(}}{{\rm{M}}_1}{\rm{),}}...{\rm{,f(}}{{\rm{M}}_D}{\rm{)])}}
\end{align}

\begin{align}
\overline {\rm{Y}} (ccr({\rm{M = }}\overline {\rm{M}} ),{\rm{X = \textbf{X}}}) = C{\rm{([f(}}{\overline {\rm{M}} _1}{\rm{),}}...{\rm{,f(}}{\overline {\rm{M}} _D}{\rm{)])}}
\end{align}

\begin{align}
{{\rm{Y}}_{effect}}{\rm{(M = \textbf{M},X = \textbf{X}) = }}{{\rm{\pounds}}_1}[\overline {\rm{Y}} - {\rm{Y}}]
\end{align}

Among them, ${{\rm{Y}}_{effect}}$ is the difference between the real prediction result and the simulated prediction result. Its value reflects the degree of influence of X on M. The influence of learned attention on the prediction result can be expressed by ${{\rm{\pounds}}_1}[\overline {\rm{Y}} - {\rm{Y}}]$. When ${{\rm{Y}}_{effect}}$ becomes smaller, it means that the attention map M learned by the network is gradually not affected by the input feature X. Note that this process only exists in the training phase.

\subsection{Loss Learning Strategy, LLS}\label{subsubsec2}

\begin{figure}
\centering 
\includegraphics[width=0.6\columnwidth]{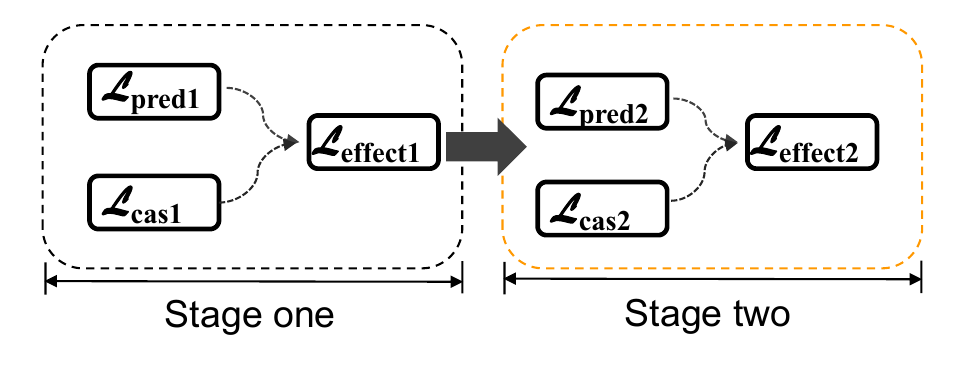}
\caption{GCAM's loss task hierarchy. The first stage is coarse-grained feature classification. The second stage builds on the first stage and the second stage classification tasks.}
\label{fig7}
\end{figure}

The CRA module primarily addresses the issue of model attention, and the network undergoes two rounds of feature extraction to filter out background information and enhance its attention to the image. However, this approach also poses challenges during training. The training process is divided into two stages, as illustrated in Fig \ref{fig7}, where the first stage serves as a precursor to the second stage. In the first stage, features are extracted from the input network before cropping for prediction, while the second stage involves prediction based on the original image cropped according to the attention map. Commonly employed loss function designs are expressed as follows (12):

\begin{align}
{\pounds_{total}} = {\pounds_{effect1}} + {\pounds_{effect2}}
\end{align}

This gives rise to a problem where, when the preceding section is poorly trained, the subsequent section's training is also adversely affected. In other words, at the outset of training, the predictions Y and ${{\rm{\bar Y}}}$ in the first phase are notably inaccurate, which can misguide the overall training process and impair performance. When the network is inadequately trained and its feature extraction capabilities are subpar, training fine-grained features on this foundation also performs poorly, resulting in significant fluctuations (i.e., instability). This, in turn, adversely affects the overall network training, leading to considerable instability, increased computational costs, and suboptimal results.

This situation gives rise to a problem: when the preceding section is inadequately trained, the subsequent section's training is likewise compromised. In other words, at the commencement of training, the predictions Y and ${{\rm{\bar Y}}}$ in the first phase are significantly inaccurate, which can mislead the overall training process and degrade performance. When the network is not well-trained and its feature extraction capabilities are subpar, training fine-grained features on this foundation also performs poorly, resulting in substantial fluctuations, denoted as ${\pounds_{total}}$. Consequently, this impacts the overall network training, leading to considerable instability, increased computational costs, and suboptimal results.

The optimization in the second stage builds upon the foundation laid in the first stage. Therefore, during the training process, it is essential to allocate a higher loss weight primarily to the first stage. As training progresses, the loss weight for the second stage should be gradually increased. Hence, we have devised a loss learning strategy to regulate the weight allocation for each task during each epoch. We employ an iteration counter, denoted as ${\alpha _i}(t)$ to control the distribution of loss weights, as expressed in the following equation (13): 

\begin{align}
\begin{array}{l}
{\pounds_{total}} = (1 - {\alpha _i}(t)){\pounds_{effect1}} + {\alpha _i}(t){\pounds_{effect2}}\\
\begin{array}{*{20}{c}}
{}&{}
\end{array}{\alpha _i}(t) \in [0,1]
\end{array}
\end{align}

In this context, ${\alpha _i}(t)$ is represented as a linear function that increases linearly with the progression of epochs. For example, with a total of 100 epochs, ${\alpha _i}(t)$ increments by epoch × 0.01 in weight with each training epoch. When designing the iteration counter, it is natural to determine parameter values based on the learning progress of each preceding task. If all prerequisite tasks have undergone a substantial number of iterations, resulting in effective training, one would anticipate a larger value for ${\alpha _i}(t)$. Conversely, if the prerequisite tasks have not received extensive training, ${\alpha _i}(t)$ should be relatively smaller. This approach draws inspiration from the observation that humans typically embark on advanced courses after completing foundational ones. Built upon this overarching design, the loss weights for each epoch can dynamically reflect the learning progress of their respective prerequisite tasks, contributing to a more stable training process.

\begin{figure}
\centering 
\includegraphics[width=0.8\columnwidth]{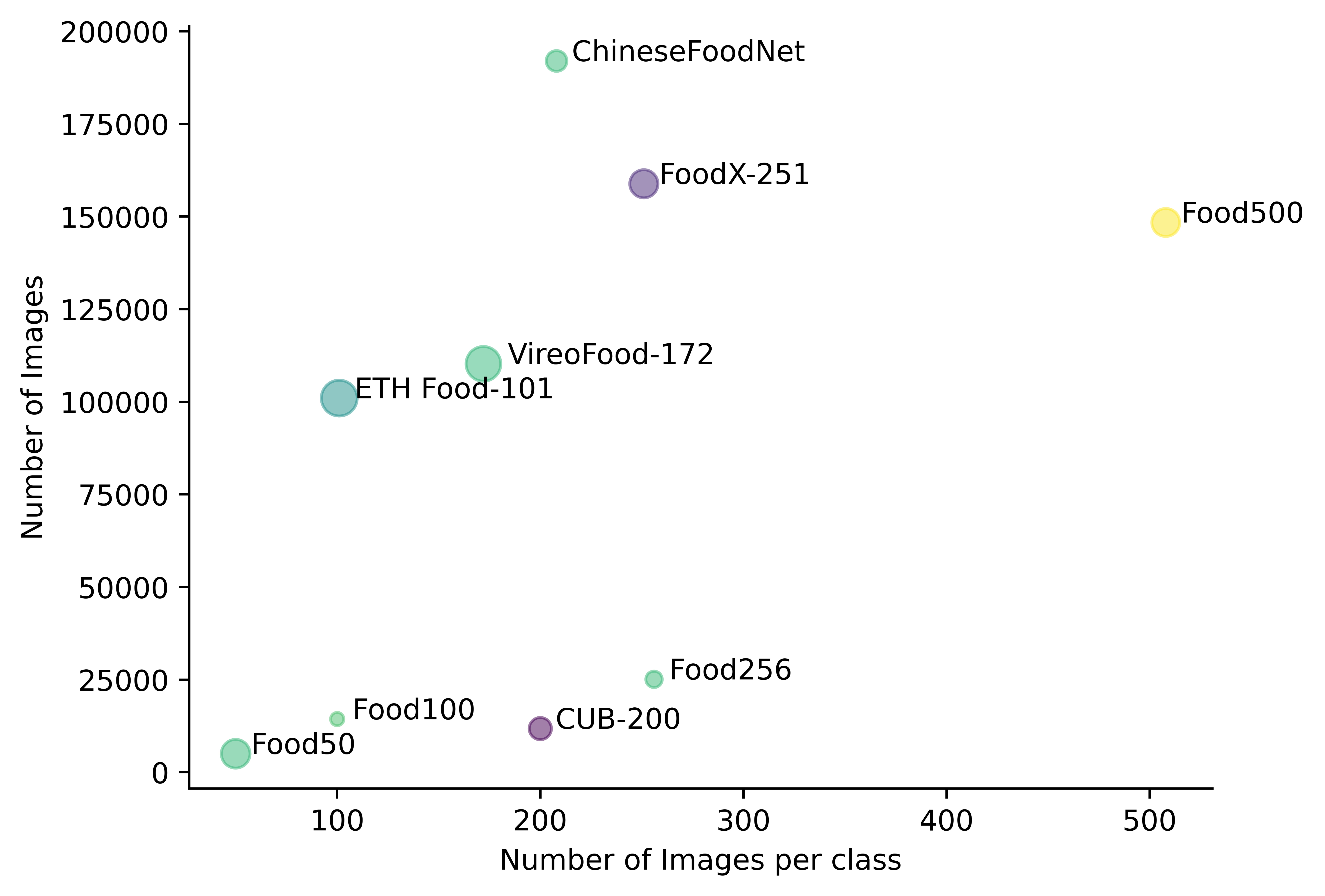}
\caption{Data distribution chart. The abscissa is the category, the ordinate is the number of data.}
\label{fig8}
\end{figure}

\section{experiment}\label{sec4}

\subsection{Setting\label{subsubsec4}}
We employed the Swim-transformer as our baseline and the cornerstone of our methodology. Swim-transformer utilizes a pre-trained model on Imagenet. The input image resolution is set to 224×224, with a feature map downsampling rate of 4. During the training phase, modeling of the model loss is performed before cropping the images, while during the testing phase, only the cropped prediction results are generated.

Model training was conducted using a batch size of 16 on a Nvidia 3090TI GPU for 160 epochs. The initial learning rate was set to 5e-2 and decayed by a factor of 0.1 starting from the 80th epoch. To enhance training stability, we applied a linear warm-up strategy for the first 5 epochs.

\subsection{Main results for the dataset\label{subsubsec4}}

\begin{table}
\normalsize
\centering
\renewcommand\arraystretch{1.3}
\scalebox{0.9}{
\begin{tabular}{cccccc} 
\toprule\multirow{2}{*}{Model}
                      & \multicolumn{2}{c}{Food101} &  \multicolumn{2}{c}{Food-172} &   \\ 

                 & Top1  & Top5                & Top1  & Top5                 &   \\ 
\toprule
AlexNet\cite{a1}            & 55.89 & -                   & 64.91 & 85.32                &   \\
Vgg16\cite{a2}              & 79.02 & 93.78               & 80.41 & 94.59                &   \\
Two-Scale CNN\cite{a3}      & 86.21 & 97.19               & 89.72 & 98.4                 &   \\
DeepFood\cite{a4}           & 77.4  & 93.7                & -     & -                    &   \\
ResNet152\cite{a5}          & 86.61 & 96.95               & 86.86 & 97.11                &   \\
Inceptionv3\cite{a6}        & 84.15 & 96.11               & 87.58 & 97.39                &   \\
DenseNet161\cite{a7}        & 86.94 & 97.03               & 86.93 & 97.17                &   \\
SENet154\cite{a8}           & 88.62 & 97.57               & 88.71 & 97.74                &   \\
WRN\cite{a9}                & 88.72 & 97.92               & -     & -                    &   \\
NTS-NET(ResNet50)\cite{a10} & 89.4  & 97.8                & 89.24 & 97.91                &   \\
WISeR[9]              & 90.27 & 98.71               & -     & -                    &   \\
PMG(ResNet50)\cite{a11}     & 86.93 & 97.21               & 89.78 & 97.45                &   \\
DCL(Resnet50)\cite{a12}     & 88.9  & 97.82               & 89.13 & 97.82                &   \\
Ours                  & \textbf{91.11} & \textbf{98.88}               & \textbf{91.03} & \textbf{98.38}                &   \\
\bottomrule
\end{tabular}}
\caption{Compare the performance on the test set with other methods. We highlight the best results in bold.}
\label{t1}
\end{table}

We will use Vireo Food172, UECFood256 and ETHFood101 food image recognition data sets for verification in this section. Fig \ref{fig8} shows the common mainstream data set category-quantity distribution chart.

The VireoFood-172 dataset\cite{d1} comprises 110,241 food images from 172 distinct categories, serving as the basis for training and testing. The ETHFood101:Food-101 dataset\cite{d2}, on the other hand, consists of 101 food categories, with 750 training images and 250 testing images per category, totaling 101,000 images. The labels for the testing images have undergone manual curation, while the training set contains some noise.

As depicted in Tab \ref{t1} of our experimental results, we compared our proposed approach to state-of-the-art methods. Overall, our method outperformed previous techniques in terms of both Top1 and Top5 accuracy. For VireoFood-172, we achieved accuracy rates of 91.03\% and 98.38\% for Top1 and Top5, respectively, representing improvements of 1.25\% and 0.56\% compared to the recently proposed PMG algorithm. For ETHFood101, our approach attained Top1 and Top5 accuracies of 91.11\% and 98.88\%, respectively, resulting in gains of 0.84\% and 0.17\% compared to the WISeR algorithm.

The UECFood256 dataset\cite{d3} is an extension of the UECFood100, encompassing 156 additional food categories. As shown in Tab \ref{t2}, we compared our method with corresponding approaches on the UECFood256 test set. It is worth noting that our proposed method yielded superior results on both the Top1 and Top5 benchmarks. Under fair conditions, our method achieved gains of 87.26\% and 96.71\% for Top1 and Top5 settings, respectively, on the UECFood256 dataset. These results unequivocally demonstrate the effectiveness of our approach.

\begin{table}
\centering
\normalsize
\renewcommand\arraystretch{1.3}
\begin{tabular}{lll} 
\toprule
Model & Top1  & Top5   \\ 
\hline
DeepFood\cite{a4}     & 63.8  & 87.2   \\
ResNet-200\cite{a5}    & 79.12 & 93.00  \\
WRN\cite{a9}           & 79.76 & 93.90  \\
WISeR\cite{a9}         & 83.15 & 95.45  \\
Two-Scale CNN\cite{a3} & 71.75 & 91.94  \\
Ours             & \textbf{87.26} & \textbf{96.71}  \\
\toprule
\end{tabular}
\caption{Comparison of the test set effects of the proposed algorithm and other algorithms on UECFood256. We highlight the best results in bold.}
\label{t2}
\end{table}

\begin{table}
\centering
\normalsize
\renewcommand\arraystretch{1.3}
\begin{tabular}{c|c} 
\toprule
Model & Accuracy(\%)  \\ 
\hline
RA-CNN\cite{a13}  & 85.3          \\
MA-CNN\cite{a14}  & 86.5          \\
WS-DAN\cite{WS-DAN} & 89.4          \\
DCL\cite{a15}     & 87.8          \\
API-Net\cite{a16} & 90            \\
CAL\cite{CAL}     & 90.6          \\
Ours    & \textbf{90.79 }        \\
\toprule
\end{tabular}
\caption{Comparison of the test set effects of the proposed algorithm and other algorithms on CUB-200. We highlight the best results in bold.}
\label{t3}
\end{table}

The California Institute of Technology-University of California, San Diego Bird-200-2011 (CUB-200-2011) dataset\cite{d4} is the most widely used dataset in fine-grained visual classification tasks. It comprises 11 subordinate categories of bird species, totaling 11,788 images, with 5,994 images allocated for training and 5,794 for testing. To demonstrate the applicability of our algorithm to other datasets, we conducted training and testing on the CUB-200 dataset, as shown in Tab \ref{t3}. Our method achieved an accuracy of 90.79\% on the CUB-200 dataset, demonstrating performance nearly on par with recent state-of-the-art methods such as WS-DAN and CAL. This indicates that our approach exhibits strong performance in other fine-grained tasks as well.

\subsection{Ablation experiment\label{subsubsec4}}

To understand the extent of improvement contributed by each component, we conducted ablation experiments on four different datasets. In these ablation experiments, aimed at examining the impact of the proposed modules on network performance, we first trained a baseline model. Subsequently, we incrementally introduced ablation targets (FDF module, CRA module, and LLS module) to this baseline. The key results are summarized in Tab \ref{t4}.

\begin{table*}
\centering 
\small
\renewcommand\arraystretch{1.5}
\setlength{\tabcolsep}{0.8mm}{
\begin{tabular}{cccccccc}
\toprule
\multirow{2}{*}{baseline} & \multirow{2}{*}{FGF} & \multirow{2}{*}{CRA} & \multirow{2}{*}{LLS} & Vireo-FOOD172 & FOOD256 & FOOD101 & CUB-200 \\
    \multicolumn{4}{c}{}& Top1/Top5   & Top1/Top5   & Top1/Top5   & Top1/Top5   \\[1ex] \hline
(a) & - & - & - & 86.06/95.70 & 83.69/95.07 & 86.54/96.90 & 84.06/92.03 \\
(b) & \textbf{$\checkmark$ } & - & - & 88.28/96.23 & 85.93/94.73 & 88.82/96.75 & 87.59/95.55 \\
(c) & - & \textbf{$\checkmark$ } & - & 87.74/96.50 & 85.59/96.61 & 87.57/97.55 & 86.22/94.25 \\
(d) & - & - & \textbf{$\checkmark$ } & 86.74/96.91 & 83.93/95.97 & 86.93/96.97 & 84.61/92.41 \\
(e) & \textbf{$\checkmark$ } & - & \textbf{$\checkmark$ } & 89.74/97.69 & 86.43/96.11 & 90.89/97.42 & 88.94/98.44 \\
(f) & - & \textbf{$\checkmark$ } & \textbf{$\checkmark$ } & 88.48/97.37 & 86.59/96.35 & 89.22/95.17 & 87.49/96.01 \\
(g) & \textbf{$\checkmark$ } & \textbf{$\checkmark$ } & \textbf{$\checkmark$ } & \textbf{91.03}/\textbf{98.38} & \textbf{87.26}/\textbf{96.71} & \textbf{91.11}/\textbf{98.83} & \textbf{90.79}/\textbf{98.52} \\ \bottomrule
\end{tabular}}
\caption{Ablation study of modules in GCAM.}
\label{t4}
\end{table*}

From Tab \ref{t4}, it is evident that the FDF module, CRA module, and LLS module all led to improvements in the model's performance. For instance, on the FOOD256 dataset, the FDF module, CRA module, and LLS module contributed 2.24\%, 1.9\%, and 0.24\% increases in Top1 accuracy, respectively, over the baseline model.

\begin{figure}
\centering 
\includegraphics[width=0.6\columnwidth]{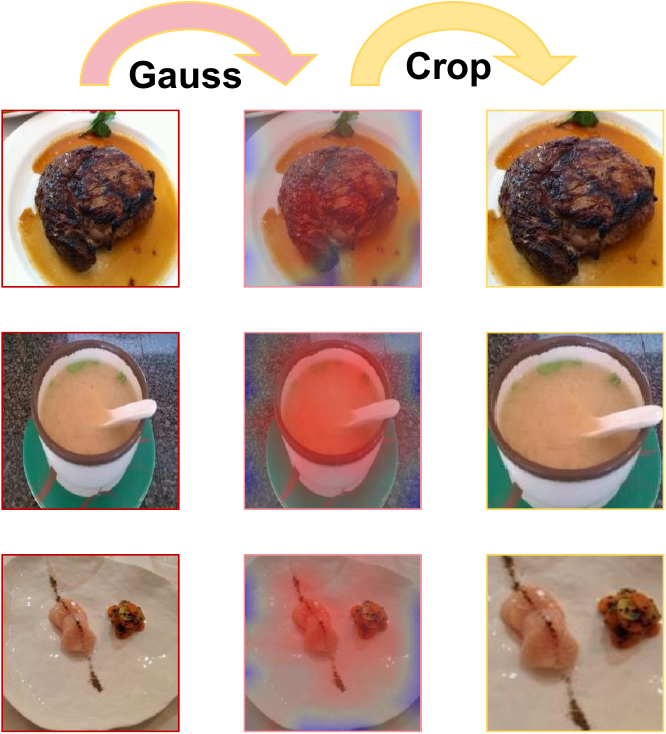}
\caption{Image cropping process from the first stage to the second stage.}
\label{fig11}
\end{figure}

Firstly, the most significant improvement is observed with the FDF module, as evidenced by the Top1 accuracy gains over the baseline for the four datasets: ETHFOOD101, UECFOOD256, Vireo-FOOD172, and CUB, which increased by 2.22\%, 2.24\%, 2.28\%, and 3.53\%, respectively. Consequently, an analysis of the FDF module reveals that during the testing process, the generated Gaussian maps are visually combined with the original images, as illustrated in Fig \ref{fig11}. Notably, the original images undergo feature extraction by the backbone, and the extracted features are used to predict the parameters ${\mu}$ and ${\sigma}$ of the Gaussian function. From the visualization in Fig \ref{fig11}, it can be observed that the regions of the Gaussian heatmaps are concentrated in areas where food items are present. This is highly advantageous for the subsequent process of transitioning from coarse to fine-grained image representation, ultimately contributing to the network's impressive predictive performance.

Next, the primary objective of the CRA module is to constrain the uncertainty in the attention mechanism through causal counterfactual reasoning, as illustrated in Tab \ref{t4}. As shown, the CRA module has led to improvements in Top1 accuracy over the baseline on the four datasets: 1.68\%, 1.9\%, 1.03\%, and 2.16\%, respectively. We observe that CRA effectively enhances both classification accuracy and attention quantification. This is because CRA serves as a module that enhances attention learning and mitigates the impact of dataset biases by quantifying the influence of facts and counterfactuals on the final predictions, thus encouraging the network to learn more effective visual attention.

Finally, in the ablation analysis of the LLS module, we observed that the addition of the LLS module to the baseline resulted in relatively modest performance improvements. The LLS module yielded an increase in Top1 accuracy of only 0.68\%, 0.24\%, 0.39\%, and 0.55\% compared to the baseline across the four datasets. We hypothesized that the LLS module was designed to balance the FDF and CRA modules. Therefore, we conducted additional ablation experiments by adding the LLS module separately to the FDF and CRA modules. As shown in Tab \ref{t4}, when combined with either the FDF or CRA module, our predictive results improved, indicating that the LLS module can dynamically reflect the learning progress of its preceding tasks based on the weight of losses at each epoch. This, in turn, enhances training stability and effectiveness.

\begin{figure}
\centering 
\includegraphics[width=0.8\columnwidth]{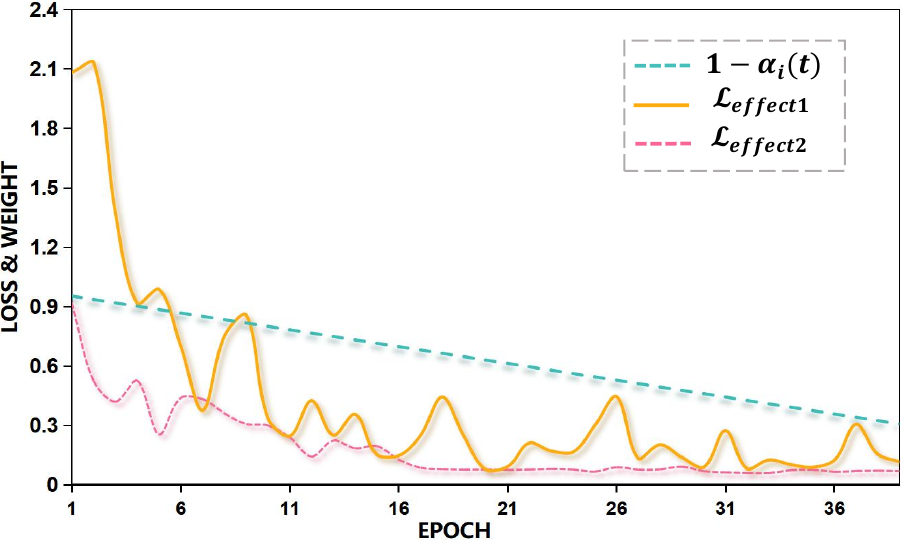}
\caption{LLS loss curve. The curve contains the loss in the first stage and the loss in the second stage, as well as the changing trend of 1-${\alpha _i}(t)$.}
\label{fig10}
\end{figure}

To illustrate the contribution of our proposed LLS strategy, we selected the first 40 epochs of training on the UECFOOD256 dataset as an example. We quantified the impact of the LLS strategy and explored the relationships between loss terms. The changing trend of loss weights during the training phase is visualized in Fig \ref{fig10}, demonstrating the effectiveness of our LLS scheme. It is evident that the weight 1- ${\alpha _i}(t)$ dynamically changes, prioritizing the preceding task with a higher weight during training. As training progresses, the loss of the preceding task decreases, and the weight is gradually shifted to the subsequent task, leading to a reduction in loss until convergence. Our method achieved excellent performance, with the main reason for outperforming the comparative methods being the hierarchical task structure introduced by our LLS model.

Overall, our module as a whole significantly improves the recognition performance of fine-grained images.

\begin{figure}
\centering 
\includegraphics[width=1.0\columnwidth]{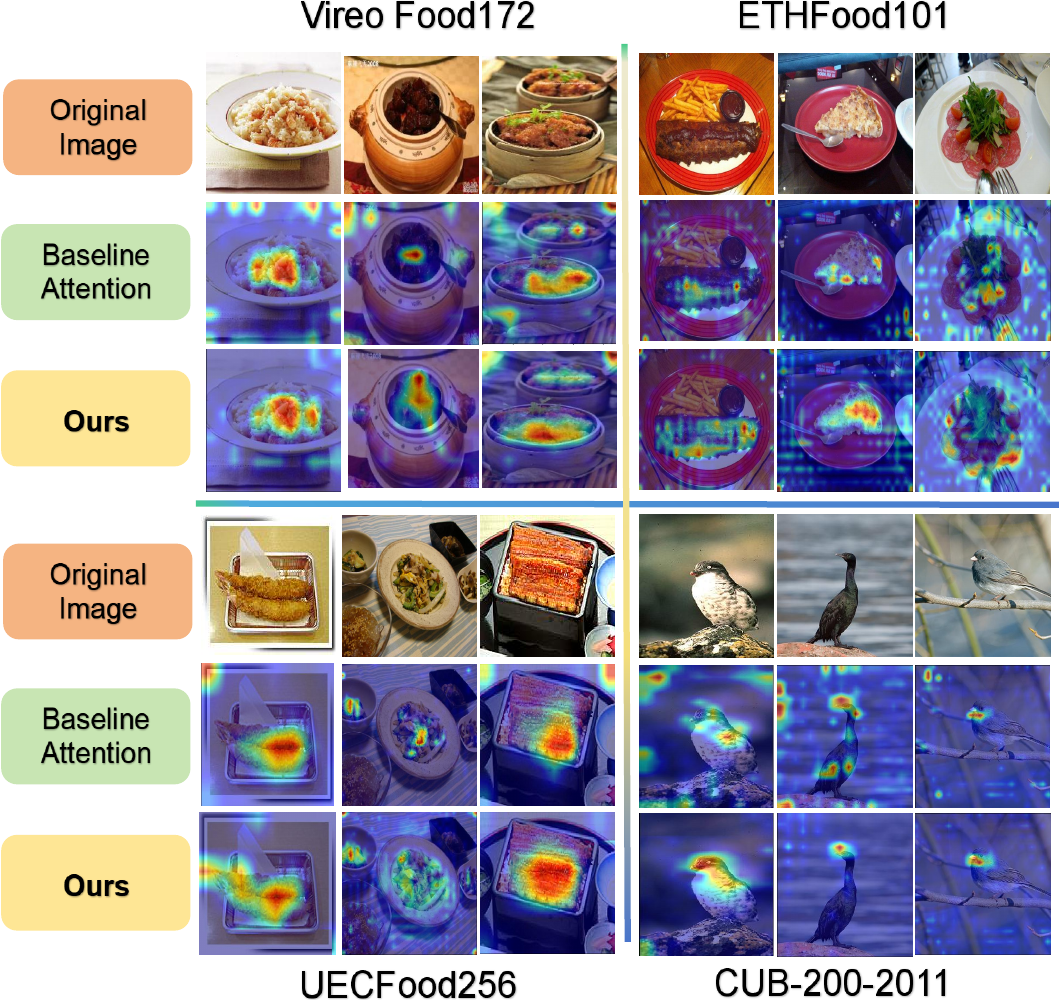}
\caption{Example of attention visualization on validation set. Randomly select an image from four data sets for visualization, including Baseline and our proposed algorithm for comparison.}
\label{fig9}
\end{figure}

\subsection{Visual result analysis\label{subsubsec4}}

To intuitively demonstrate the results and highlight the effectiveness of our proposed model, Fig \ref{fig9} presents visualizations of the GCAM model across four datasets. We have included visualizations for the original images, the Baseline model, and the GCAM model for comparison. In comparison to the Baseline model, our proposed GCAM model exhibits a greater focus on the target region's features and excels in finer details of object recognition, especially in scenarios involving intricate object distinctions.

For instance, in the second column of examples from the Vireo Food172 dataset and the CUB-200 dataset, under complex background conditions, the Baseline model tends to extract features influenced by background information, making it challenging for the network to concentrate its attention on the object region, resulting in inferior recognition performance. In contrast, the GCAM model effectively emphasizes fine-grained object information, accurately pinpointing the target area and capturing the nuanced features of the object. This gives more weight to the object, significantly reducing interference from complex backgrounds and thereby enhancing network performance.

\section{Conclusion}

In this paper, we introduce a novel Gaussian and causal-attention model(GCAM), which enhances network focus on object locations by training a Gaussian distribution map of object positions and jointly weighting it with global feature inputs. Additionally, we construct a causal graph and loss function by comparing facts and counterfactuals to quantify the quality improvement of the network's attention mechanism and reduce its susceptibility to data drift, ensuring that attention is not solely directed toward partial objects.

Furthermore, we design a learnable loss strategy, LLS, to mitigate training instability in the initial phases. Extensive experiments validate the superior performance of the proposed algorithm and the effectiveness of its constituent modules. In the future, we aim to extend our research to multi-label classification, achieving effective detection and recognition in scenarios involving diverse objects within a single image.

\section{Acknowledgments}

No funding was received to assist with the preparation of this manuscript.

\section{Conflict of Interests}
 The authors declare that they have no conflict of
interest.

\section{Data availability and access}

The data that support the finding of this study are openly available. The Vireo-FOOD172 data source, DOI: http://dx.doi.org/10.1145/2964284.2964315; ETHFood101 data source, DOI: https://doi.org/10.1007 /978-3-319-10599-4\_29; UECFood256 data source, DOI: https://doi.org/10.1007/978-3-319-16199-0\_1; CUB-200 data comes from Caltech Vision Lab open source link is https://www.vision.caltech.edu/datasets/cub\_200\_2011


\section*{Declarations}


\begin{itemize}
\item Funding

No funding was received to assist with the preparation of this manuscript.

\item Competing Interests

The authors declare that they have no competing interests

\item Authors' contributions

All authors contributed to the study conception and design. Material preparation, data analysis were performed by [Yue Hu], [Jiazhan Gao] and [Tianxing Yang], and Exprenment was performed by [Guohang Zhuang]. The first draft of the manuscript was written by [Guohang Zhuang] and all authors commented on previous versions of the manuscript. All authors read and approved the final manuscript.

\item Ethical and informed consent for data used

This article received written informed consent from all authors


\end{itemize}

\bibliographystyle{unsrt}
\bibliography{sn-bibliography.bib}


\end{document}